# Continuous Time Markov Networks

**Tal El-Hay**
School of Computer Science
The Hebrew University
{tale,nir}@cs.huji.ac.il

**Nir Friedman**

**Daphne Koller**
Department of Computer Science
Stanford University
koller@cs.stanford.edu

**Raz Kupferman**
Institute of Mathematics
The Hebrew University
raz@math.huji.ac.il

## Abstract

A central task in many applications is reasoning about processes that change over continuous time. Recently, Nodelman et al. introduced *continuous time Bayesian networks (CTBNs)*, a structured representation for representing *Continuous Time Markov Processes* over a structured state space. In this paper, we introduce *continuous time Markov networks (CTMNs)*, an alternative representation language that represents a different type of continuous-time dynamics, particularly appropriate for modeling biological and chemical systems. In this language, the dynamics of the process is described as an interplay between two forces: the tendency of each entity to change its state, which we model using a continuous-time *proposal process* that suggests possible local changes to the state of the system at different rates; and a global *fitness* or *energy* function of the entire system, governing the probability that a proposed change is accepted, which we capture by a Markov network that encodes the fitness of different states. We show that the fitness distribution is also the stationary distribution of the Markov process, so that this representation provides a characterization of a temporal process whose stationary distribution has a compact graphical representation. We describe the semantics of the representation, its basic properties, and how it compares to CTBNs. We also provide an algorithm for learning such models from data, and demonstrate its potential benefit over other learning approaches.

## 1 Introduction

In many applications, we reason about processes that evolve over time. Such processes can involve short time scales (e.g., the dynamics of molecules) or very long ones (e.g., evolution). In both examples, there is no obvious discrete "time unit" by which the process evolves. Rather, it is more natural to view the process as changing in a continuous time: the system is in some state for a certain duration, and then transitions to another state. The language of *continuous time Markov processes* (CTMPs) provides an elegant mathematical framework to reason about the probability of trajectories of such systems. Unfortunately, when we consider a system with multiple components, this representation grows exponentially in the number of components. Thus, we aim to construct a representation language for CTMPs that can compactly encode natural processes over high-dimensional state spaces. Importantly, the representation should also facilitate effective inference and learning.

Recently, Nodelman et al. [8, 9, 10, 11] introduced the representation language of *continuous time Bayesian networks* (CTBNs), which provides a factorized, component-based representation of CTMP: each component is characterized by a conditional CTMP dynamics, which describes its local evolution as a function of the current state of its parents in the network. This representation is natural for describing systems with a sparse structure of local influences between components. Nodelman et al. provide algorithms for efficient approximate inference in CTBNs, and for learning them from both complete and incomplete data.

In this paper, we introduce *continuous time Markov networks*, which have a different representational bias. Our motivating example is modeling the evolution of biological sequences such as proteins. In this example, the state of the system at any given time is the sequence of amino acids encoded by the gene of interest. As evolution progresses, the sequence is continually modified by local mutations that change individual amino acids. The mutations for different amino acids occur independently, but the probability that these local mutations survive depends on global aspects of the new sequence. For example, a mutation may be accepted only if the new sequence of amino acids folds properly into a functional protein, which occurs only if pairs of amino acids that are in contact with each other in the folded protein have complementary charges. Thus, although the modifications are local, global constraints on the protein structure and function introduce dependencies.

To capture such situations, we introduce a representation where we specify the dynamics of the process using two components. The first is a *proposal process* that attempts to change individual components of the system. In our example, this process will determine the rate of random mutations in protein sequences. The second is an equilibrium distribution, which encodes preferences over global configurations of the system. In our example, an approximation of the fitness of the folded protein. The equilibrium distribution is a static quantity that encodes preferences among states of the system, rather than dynamics of changes. The actual dynamics of the system are determined by the interplay between these two forces: local mutations and global fitness. We represent the equilibrium distribution compactly using a Markov network, or, more generally, a feature-based log-linear model.

Importantly, as we shall see, the equilibrium distribution parameter is indeed the *equilibrium* distribution of the process. Thus, our representation provides an explicit representation of both the dynamics of the system and its asymptotic limit. Moreover, this representation ensures that the equilibrium distribution has a pre-specified simple structure. Thus, we can view our framework as a *continuous-time Markov network* (CTMN) — a Markov network that evolves over continuous time. From a different perspective, our representation allows us to capture a family of temporal processes whose stationary distribution has a certain locality structure. Such processes occur often in biological and physical systems. For example, recent results of Socolich et al. [13] suggest that pairwise Markov networks can fairly accurately capture the fitness of protein sequences.

We provide a reduction from CTMNs to CTBNs, allowing us to use CTBN algorithms [7, 11] to perform effective approximate inference in CTMNs. More importantly, we also provide a procedure for learning CTMN parameters from data. This procedure allows us to estimate the stationary distribution from observations of the system's dynamics. This is important in applications where the stationary distribution provides insight about the domain of application. In the protein evolution example, the stationary distribution provides a description of the evolutionary forces that shape the protein and thus gives important clues about protein structure and function.

## 2 Reversible Continuous Time Markov Processes

We now briefly summarize the relevant properties of continuous time Markov processes that will be needed below. We refer the interested reader to Taylor and Karlin [14] and Chung [2] for more thorough expositions. Suppose we have a family of random variables $\{\boldsymbol{X}(t) : t \geq 0\}$ where the continuous index $t$ denotes time. A joint distribution over these random variables is a homogeneous *continuous time Markov process* (CTMP) if it satisfies the *Markov property*

$$\Pr(\boldsymbol{X}(t_{k+1})|\boldsymbol{X}(t_k),\ldots,\boldsymbol{X}(t_0)) = \Pr(\boldsymbol{X}(t_{k+1})|\boldsymbol{X}(t_k))$$

for all $t_{k+1} > t_k > \ldots > t_0$, and time-homogeneity,

$$\Pr(\boldsymbol{X}(s+t) = \boldsymbol{y}|\boldsymbol{X}(s) = \boldsymbol{x}) = \\ \Pr(\boldsymbol{X}(s'+t) = \boldsymbol{y}|\boldsymbol{X}(s') = \boldsymbol{x})$$

for all $s, s'$ and $t > 0$.

The dynamics of a CTMP are fully determined by the *Markov transition function*,

$$p_{\boldsymbol{x},\boldsymbol{y}}(t) = \Pr(\boldsymbol{X}(s+t) = \boldsymbol{y}|\boldsymbol{X}(s) = \boldsymbol{x}),$$

where time-homogeneity implies that the right hand side does not depend on $s$. Provided that the transition function satisfies certain analytical properties (see [2]) the dynamics are fully captured by a constant matrix $\boldsymbol{Q}$ — the *rate*, or *intensity matrix* — whose entries $q_{\boldsymbol{x},\boldsymbol{y}}$ are defined by

$$q_{\boldsymbol{x},\boldsymbol{y}} = \lim_{h \downarrow 0} \frac{p_{\boldsymbol{x},\boldsymbol{y}}(h) - \boldsymbol{1}\{\boldsymbol{x} = \boldsymbol{y}\}}{h}, \tag{1}$$

where $\boldsymbol{1}\{\}$ is the *indicator function* which takes the value 1 when the condition in the argument holds and 0 otherwise. The Markov process can also be viewed as a generative process: The process starts in some state $\boldsymbol{x}$. After spending a finite amount of time at $\boldsymbol{x}$, it transitions, at a random time, to a random state $\boldsymbol{y} \neq \boldsymbol{x}$. The transition times to the various states are exponentially distributed, with rate parameters $q_{\boldsymbol{x},\boldsymbol{y}}$. The diagonal elements of $\boldsymbol{Q}$ are set to ensure the constraint that each row sums up to zero.

If the process satisfies certain conditions (reachability) then the limit

$$\boldsymbol{\pi}_{\boldsymbol{x}} = \lim_{t \to \infty} p_{\boldsymbol{y},\boldsymbol{x}}(t)$$

exists and is independent of the initial state $\boldsymbol{y}$. That is, in the long time limit, the probability of visiting state $\boldsymbol{x}$ is independent of the initial state at time 0. The distribution $\boldsymbol{\pi}_{\boldsymbol{x}}$ is called the *stationary distribution* of the process. A CTMP is called *stationary* if $P(\boldsymbol{X}(0) = \boldsymbol{x}) = \boldsymbol{\pi}_{\boldsymbol{x}}$, that is, if the initial state is sampled from the stationary distribution. A stationary CTMP is called *reversible* if for every $\boldsymbol{x}, \boldsymbol{y}$, and $t > 0$

$$\Pr(X(t) = \boldsymbol{y}|X(0) = \boldsymbol{x}) = \Pr(X(0) = \boldsymbol{y}|X(t) = \boldsymbol{x}).$$

This condition implies that the process is statistically equivalent to itself running backward in time. Reversibility is intrinsic to many physical systems where the microscopic dynamics are time-reversible. Reversibility can be formulated as a property on the Markov transition function, where for every $\boldsymbol{x}, \boldsymbol{y}$, and $t > 0$

$$\boldsymbol{\pi}_{\boldsymbol{x}} p_{\boldsymbol{x},\boldsymbol{y}}(t) = \boldsymbol{\pi}_{\boldsymbol{y}} p_{\boldsymbol{y},\boldsymbol{x}}(t).$$

This identity is known as the *detailed balance* condition. To better understand the constraint, we can examine the implications of reversibility on the rate matrix $\boldsymbol{Q}$.

**Proposition 2.1:** *A CTMP is reversible if and only if its rate matrix can be expressed as*

$$q_{x,y} = \pi_y s_{x,y},$$

*where $s_{x,y}$ are the entries of a symmetric matrix (that is, $s_{x,y} = s_{y,x}$).*

In other words, in a reversible CTMP, the asymmetry in transition rates can be interpreted as resulting entirely from preferences of the stationary distribution.

## 3 Continuous Time Metropolis Processes

We start by considering a reformulation of reversible CTMPs as a continuous time version of the Metropolis sampling process. We view the process as an interplay between two factors. The first is an unbiased random process that attempts to transition between states of the system, and the second is the tendency of the system to remain in more probable states. This latter probability is taken to be the stationary distribution of the process. The structure of the process can be thought of as going through iterations of proposed transitions that are either accepted or rejected, similar to the Metropolis sampler [6].

To formally describe such a process, we need to describe these two components. The first is the unbiased proposal of transitions. These proposals occur at fixed rates. We denote by $r_{x,y}$ the rate at which proposals to transition $x \to y$ occur. This in effect defines a CTMP process with rate matrix $R$. To ensure an unbiased proposal, we require $R$ to be symmetric. (The stationary distribution of a symmetric rate matrix is the uniform distribution.)

The second component is a decision whether to accept or reject the proposed transition. The decision whether to accept the transition $x \to y$ depends on the probability ratio of these states at equilibrium. We assume that we are given a target distribution, which should coincide with the equilibrium distribution $\pi$. As we shall see, to reach the target equilibrium distribution, the acceptance probability should satisfy a simple condition. To make this precise, we assume we have an *acceptance function* $f$ that takes as an argument the ratio $\pi_y/\pi_x$ and returns the probability of accepting transition $x \to y$. This function should return a value between 0 and 1, and satisfy the functional relation

$$f(z) = zf\left(\frac{1}{z}\right). \tag{2}$$

Two functions that satisfy these conditions are

$$\begin{aligned} f_{\text{Metropolis}}(z) &= \min(1, z) \\ f_{\text{logistic}}(z) &= \frac{1}{1 + \frac{1}{z}}. \end{aligned}$$

The function $f_{\text{Metropolis}}$ is the standard one used in Metropolis sampling. The function $f_{\text{logistic}}$ is closely linked to logistic regression. It is continuously differentiable, which, as we shall see, facilitates the subsequent analysis.

Formally, a *continuous time Metropolis process* is defined by a symmetric matrix $R$, a distribution $\pi$, and a real-valued function $f$. The semantics of the process are defined in a generative manner. Starting at an initial state $x$, the system remains in the state until receiving a proposed transition $x \to y$ with rate $r_{x,y}$. This proposal is then accepted with probability $f(\pi_y/\pi_x)$. If it is accepted, the system transitions to state $y$; otherwise it remains in state $x$. This process is repeated indefinitely.

To formulate the statistical dynamics of the system, consider a short time interval $h$. In this case, the probability of a proposal of the transition $x \to y$ is roughly $h \cdot r_{x,y}$. Since the proposed transition is accepted with probability $f(\pi_y/\pi_x)$, we have:

$$p_{x,y}(h) \approx h \cdot r_{x,y} \cdot f\left(\frac{\pi_y}{\pi_x}\right).$$

Plugging this into Eq. (1) we conclude that the off-diagonal elements of $Q$ are

$$q_{x,y} = r_{x,y} \cdot f\left(\frac{\pi_y}{\pi_x}\right). \tag{3}$$

**Proposition 3.1:** *Consider a continuous time Metropolis process defined as in Eq. (3). Then, this CTMP is reversible, and its stationary distribution is $\pi$.*

**Proof:** The condition on $f$ implies that

$$\frac{1}{\pi_y}f\left(\frac{\pi_y}{\pi_x}\right) = \frac{1}{\pi_x}f\left(\frac{\pi_x}{\pi_y}\right),$$

Thus, it follows that $q_{x,y}$ is of the form $s_{x,y}\pi_y$, i.e., that the process is reversible. Moreover, it implies that the stationary distribution of the process is $\pi$. ∎

The inverse result is also easy to obtain.

**Proposition 3.2:** *Any reversible CTMP can be represented as a continuous time Metropolis process.*

**Proof:** According to Proposition 2.1 we can write $q_{x,y} = \pi_y s_{x,y}$ for a symmetric matrix $s_{x,y}$. Define

$$r_{x,y} = s_{x,y} \frac{\pi_y}{f\left(\frac{\pi_y}{\pi_x}\right)},$$

so that $q_{x,y} = r_{x,y} \cdot f\left(\frac{\pi_y}{\pi_x}\right)$. Together, $s_{x,y} = s_{y,x}$ and Eq. (2) imply that $r_{x,y} = r_{y,x}$. Thus, $R$ is symmetric and together with $\pi$ defines a continuous time Metropolis process which is equivalent to the original reversible CTMP. ∎

We conclude that continuous time Metropolis processes are a general reparameterization of reversible CTMPs.

## 4 Continuous Time Markov Networks

We are interested in dealing with structured, multi-component systems, whose state description can be viewed as an assignment to some set of state variables $\boldsymbol{X} = \langle X_1, X_2, \ldots, X_n \rangle$, where each $X_i$ assumes a finite set of values. The main challenge is dealing with the large state space (exponential in $n$). We aim to find succinct representations of the system's dynamics within the framework of continuous time Metropolis processes. We do so in two stages, first dealing with the proposal rate matrix $\boldsymbol{R}$, and then with the equilibrium distribution $\boldsymbol{\pi}$.

Our first assumption is that proposed transitions are local. Specifically, we require that, for $\boldsymbol{x} \neq \boldsymbol{y}$

$$r_{\boldsymbol{x},\boldsymbol{y}} = \begin{cases} r^i_{x_i,y_i} & (x_j = y_j) \; \forall j \neq i \\ 0 & \text{otherwise} \end{cases} \quad (4)$$

where $\boldsymbol{R}^i = \{r^i_{x_i,y_i}\}$ are symmetric local transition rates for $X_i$. Thus, we allow only one component to change at a time and the proposal rates do not depend on the global state of the system.

The second assumption concerns the structure of the stationary distribution $\boldsymbol{\pi}$. *Log-linear models* or *Markov networks* provide a general framework to describe structured distributions. A log-linear model is described by a set of *features*, each one encoding a local property of the system that involves few variables. For example, the function $\boldsymbol{1}\{X_1 = X_2\}$ is a feature that only involves two variables.

A feature-based Markov network is defined by a vector of features, $\boldsymbol{s} = \langle s_1, \ldots, s_K \rangle$, where each feature $s_k$ assigns a real number to the state of the system. We further assume that each feature $s_k$ is a function of a (usually small) subset $\boldsymbol{D}_k \subseteq \boldsymbol{X}$ of variables. We use the notation $\boldsymbol{x}|_{\boldsymbol{D}_k}$ to denote the projection of $\boldsymbol{x}$ on the subset of variables $\boldsymbol{D}_k$. Thus, $s_k$ is a function of $\boldsymbol{x}|_{\boldsymbol{D}_k}$; however, for notational convenience, we sometimes use $s_k(\boldsymbol{x})$ as a shorthand for $s_k(\boldsymbol{x}|_{\boldsymbol{D}_k})$.

Based on a set of features, we define a distribution by assigning different weights to each feature. These weights represent the relative importance of each feature. We use the notation $\boldsymbol{\theta} = \langle \theta_1, \ldots, \theta_K \rangle \in \mathbb{R}^K$ to denote the vector of weights or *parameters*. The equilibrium distribution represented by $\boldsymbol{s}$ and $\boldsymbol{\theta}$ takes the log-linear form

$$\boldsymbol{\pi}_{\boldsymbol{x}} = \frac{1}{Z(\boldsymbol{\theta})} \exp\left\{ \sum_k \theta_k \cdot s_k(\boldsymbol{x}|_{\boldsymbol{D}_k}) \right\}, \quad (5)$$

where the *partition function* $Z(\boldsymbol{\theta})$ is the normalizing factor.

The structure of the equilibrium distribution can be represented as an undirected graph $\mathcal{G}$ — the nodes of $\mathcal{G}$ represent the variables $\{X_1, \ldots, X_n\}$. If $X_i, X_j \in \boldsymbol{D}_k$ for some $k$, then there is an edge between the corresponding nodes. Thus, for every feature $s_k$, the nodes that represent the variables in $\boldsymbol{D}_k$ form a clique in the graph $\mathcal{G}$. We define the *Markov Blanket*, $\mathcal{N}_{\mathcal{G}}(i)$, of the variable $X_i$ as the set of neighbors of $X_i$ in the graph $\mathcal{G}$ [12].

**Example 4.1 :** Consider a four-variable process $\{X_1, X_2, X_3, X_4\}$, where each variable takes binary

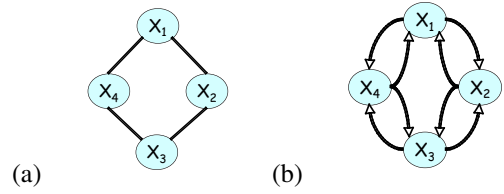

(a) (b)

Figure 1: (a) The Markov network structure for Example 4.1. (b) The corresponding CTBN structure.

values, with the following set of features:

$s_1(X_1) = \boldsymbol{1}\{X_1 = 1\} \quad s_5(X_1, X_2) = \boldsymbol{1}\{X_1 = X_2\}$
$s_2(X_2) = \boldsymbol{1}\{X_2 = 1\} \quad s_6(X_2, X_3) = \boldsymbol{1}\{X_2 = X_3\}$
$s_3(X_3) = \boldsymbol{1}\{X_3 = 1\} \quad s_7(X_3, X_4) = \boldsymbol{1}\{X_3 = X_4\}$
$s_4(X_4) = \boldsymbol{1}\{X_4 = 1\} \quad s_8(X_1, X_4) = \boldsymbol{1}\{X_1 = X_4\}$

Note that all these features involve at most two variables. The corresponding graph structure is shown in Figure 1(a). In this example $\mathcal{N}(1) = \{X_2, X_4\}$, $\mathcal{N}(2) = \{X_1, X_3\}$, $\mathcal{N}(3) = \{X_2, X_4\}$, and $\mathcal{N}(4) = \{X_1, X_3\}$. ∎

We now take advantage of the structured representation of both $\boldsymbol{R}$ and $\boldsymbol{\pi}$ to get a more succinct representation of the rate matrix $\boldsymbol{Q}$ of the process. We exploit the facts that $\boldsymbol{\pi}$ appears explicitly in the rate only as a ratio $\boldsymbol{\pi}_{\boldsymbol{y}}/\boldsymbol{\pi}_{\boldsymbol{x}}$, and moreover that the proposal process includes only transitions that modify a single variable. Thus, we only examine ratios where $\boldsymbol{y}$ and $\boldsymbol{x}$ agree on all variables but one. It is straightforward to show that if $\boldsymbol{x}$ and $\boldsymbol{y}$ are two states that are identical except for the value of $X_i$ and $\boldsymbol{u}_i = \boldsymbol{x}|_{\mathcal{N}(i)}$, then

$$\boldsymbol{\pi}_{\boldsymbol{y}}/\boldsymbol{\pi}_{\boldsymbol{x}} = g_i(x_i \to y_i | \boldsymbol{u}_i),$$

where

$$g_i(x_i \to y_i | \boldsymbol{u}_i) = \exp\left\{ \sum_{k: X_i \in \boldsymbol{D}_k} \theta_k [s_k(y_i, \boldsymbol{u}_i) - s_k(x_i, \boldsymbol{u}_i)] \right\}.$$

Note that, if $X_i \in \boldsymbol{D}_k$, then $\boldsymbol{D}_k \subseteq \mathcal{N}(i) \cup \{X_i\}$. Thus, the function $g_i$ is well defined.

Thus, the acceptance probability of a change in $X_i$ depends only on the state of variables in its Markov blanket. This property is heavily used for Gibbs sampling in Markov networks. Depending on the choice of features, these dependencies can be very sparse, or involve all the variables in the process.

To summarize, assuming a local form for $\boldsymbol{R}$ and a log-linear form for $\boldsymbol{\pi}$, we can further simplify the definition of the rate matrix $\boldsymbol{Q}$. If $\boldsymbol{x}$ and $\boldsymbol{y}$ are two states that differ only in the $i$'th variable, then

$$q_{\boldsymbol{x},\boldsymbol{y}} = r^i_{x_i,y_i} f(g_i(x_i \to y_i | \boldsymbol{u}_i)), \quad (6)$$

where $\boldsymbol{u}_i = \boldsymbol{x}|_{\mathcal{N}(i)}$. All other off-diagonal entries are 0, and the diagonal entries are set to ensure that the sum of

each row is 0. We call a process with a $Q$ matrix of the form Eq. (6) a *Continuous time Markov Network* (CTMN).

One consequence of the form of the CTMN rate matrix Eq. (6) is that the dynamics of the $i$'th variable depend directly only on the dynamics of its neighbors. As we can expect, we can use this property to discuss independencies among variables in the network. However, since we are examining a continuous process, we need to consider independencies between full trajectories (see also [8]).

**Theorem 4.2:** Consider a CTMN with a stationary distribution represented by a graph $\mathcal{G}$. If $A, B, C$ are subsets of $X$ such that $C$ separates $A$ from $B$ in $\mathcal{G}$, then the trajectories of $A$ and $B$ are conditionally independent, given observation of the full trajectory of $C$.

**Proof:** (sketch) Using the global independence properties of a Markov network (see for example, [12]), we have that $\pi$ can be written as a product of two function each with its own domain $X_1$ and $X_2$ such that $X_1 \cap X_2 = C$ and $A \subseteq X_1$ and $B \subseteq X_2$. Once the trajectories of variables in $C$ are given, the dynamics of variables in $X_1 - C$ and $X_2 - C$ are two independent CTMNs, each with its own stationary distribution. As a consequence we get the desired independence. ∎

That is, the usual conditional separation criterion in Markov networks [12] applies in a trajectory-wise fashion to CTMNs.

It is important to note that although we can represent any reversible CTMP as a continuous time Metropolis process, once we move to CTMNs this is no longer the case. The main restriction is that, in CTMNs as we have defined them, each transition involves a change in the state of exactly one component. Thus, although the language of Markov networks allow to describe arbitrary equilibrium distributions (potentially with an exponential number of features), the restrictions on $R$ limit the range of processes we can describe as CTMNs. As an example of a domain where CTMNs are not suitable, consider reasoning about biochemical systems, where each component of the state is the number of molecules of a particular species and transitions correspond to chemical reactions. For example, a reaction might be one that takes an $OH$ molecule and an $H$ molecule and replace them by an $H_2O$ molecule. If reactions are reversible (i.e., $H_2O$ can break into $OH$ and $H$ molecules), then this process might be described by a reversible CTMP. However, since reactions change several components at once, we cannot describe such system as a CTMN.

## 5 Connection to CTBNs

The factored form of Eq. (6) allows us to relate CTMNs with CTBNs. A CTBN is defined by a *directed* (often cyclic) graph whose nodes correspond to variables of the process, and whose edges represent direct influences of one variable on the evolution of another. More precisely, a CTBN is defined by a collection of *conditional rate matrices* (also called conditional intensity matrices). For each $X_i$, and for each possible value $u_i$ of its direct parents in the CTBN graph, the matrix $Q^{X_i|u_i}$ is a rate matrix over the state space of $X_i$. These conditional rate matrices are combined into a global rate matrix by a process Nodelman et al. [9] call amalgamation. Briefly, if $x$ and $y$ are identical except for the value of $X_i$, then

$$q_{x,y} = q_{x_i,y_i}^{X_i|u_i} \qquad (7)$$

where $u_i = x|_{\mathbf{Pa}_i}$ is the assignment to $X_i$'s parents in the state $x$. That is, the rate of transition from $x$ to $y$ is the conditional rate of $X_i$ changing from $x_i$ to $y_i$ given the state of its parents. Again, all other off-diagonal elements, where more than one variable changes, are set to 0.

This form is similar to the rate matrix of CTMNs shown in Eq. (6). Thus, given a CTMN, we can build an equivalent CTBN by setting the parents of each $X_i$ to be $\mathcal{N}(i)$, and using the conditional rates:

$$q_{x_i,y_i}^{X_i|u_i} = r_{x_i,y_i}^i g_i(x_i \to y_i \mid x|_{\mathcal{N}(i)}) \qquad (8)$$

Figure 1(b) shows the CTBN structures corresponding to the CTMN of Example 4.1. In general, the CTBN graph corresponding to a given CTMN is built by replacing each undirected arc by a pair of directed ones. This matches the intuition that if $X_i$ and $X_j$ appear in the context of some feature, then they mutually influence each other.

As this transformation shows, the class of processes that can be encoded using CTMNs is a subclass of CTBNs. In a sense, this is not surprising, as a CTBN can encode any Markov process where at most one variable can transition at a time. However, the CTMN representation imposes a particular parametrization of the system dynamics in terms of the local proposal process and the global equilibrium distribution. This parametrization violates both local and global parameter independence [5] in the resulting CTBN. In particular, a transition between $x_i$ and $y_i$ is proposed at the same rate, regardless of whether it is globally advantageous (in terms of equilibrium preferences). As we shall see, this property is important for our ability to effectively estimate these rate parameters.

Moreover, as we have seen, this parametrization guarantees that the stationary distribution of the process factorizes as a particular Markov network. In general, even a fairly sparse CTBN gives rise to a fully entangled stationary distribution that cannot be factorized. Indeed, even computing the stationary distribution of a given CTBN is a hard computational problem. By contrast, we have defined a model of temporal dynamics that gives rise to a natural and interpretable form for the stationary distribution. This property is critical in applications where the stationary distribution is the key element in understanding the system.

Yet, the ability to transform a CTMN into a CTBN allows us to harness the recently developed approximate in-ference methods for CTBNs [11, 7], including for the E-step used when learning CTMNs for partially observable data.

## 6 Parameter Learning

We now consider the problem of learning the parametrization of CTMNs from data. Thus, we assume we are given the form of $\boldsymbol{\pi}$, that is, the set of features $\boldsymbol{s}$, and need to learn the parameters $\boldsymbol{\theta}$ governing $\boldsymbol{\pi}$ and the local rate matrices $\boldsymbol{R}^i$ that govern the proposal rates for each variable. We start by considering this problem in the context of *complete data*, where our observations consist of full trajectories of the system. As we show, we define a gradient ascent procedure for learning the parameters from such data.

This result also enables us to learn from incomplete data using the standard EM procedure. Namely, we can use existing CTBNs inference algorithms to perform the E-step effectively when learning from partially observable data to compute expected sufficient statistics. The M-step is then an application of the learning procedure for complete data with these expected sufficient statistics. This combination is quite standard and follows the lines of similar procedure for CTBNs [10], and therefore we do not expand on it here.

### 6.1 The Likelihood Function

A key concept in addressing the learning problem is the likelihood function, which determines how the probability of the observations depends on the parameters.

We assume that the data is complete, and thus our observations consist of a trajectory of the system that can be described as a sequence of intervals, where in each interval the system is in one state. Using the relationship to CTBNs, we can use the results of Nodelman *et al.* [9] to write the probability of the data as a function of sufficient statistics and entries in the conditional rate matrices of Eq. (8). A problem with this approach is that the entries in the conditional rate matrix involve both parameters from $\boldsymbol{R}^i$ and parameters from $\boldsymbol{\theta}$. Thus, the resulting likelihood function couples the estimation of these two sets of parameters.

However, if we had additional information, we could decouple these two sets of parameters. Suppose we observe not only the actual trajectories, but also the rejected proposals; see Figure 2. With this additional information, we can estimate the rate of different proposals, independently of whether they were accepted or not. Similarly, we can estimate the equilibrium distribution from the accepted and rejected proposals. Thus, we are going to view our learning problem as a partial data problem where the annotation of rejected proposals is the missing data.

To formalize these ideas, assume that our evidence is a trajectory annotated with proposal attempts. We describe such a trajectory using three vectors; see Figure 2. The first vector, $\boldsymbol{\tau} = \langle \tau[1], \ldots, \tau[M+1] \rangle$, represents the time intervals between consecutive proposals. Thus, the first pro-posal took place at time $\tau[1]$, the second at time $\tau[1] + \tau[2]$, and so on. The last entry in this vector is the time between the last proposal and the end of the observed time interval. The second vector, $\Xi = \langle \boldsymbol{x}[0], \boldsymbol{x}[1], \ldots, \boldsymbol{x}[M] \rangle$, denotes the actual state of the system after each proposal was made. Thus, $\boldsymbol{x}[0]$ is the initial state of the system, $\boldsymbol{x}[1]$ is the state after the first proposal, and so on. Finally, $\Upsilon = \langle \boldsymbol{y}[1], \ldots, \boldsymbol{y}[M] \rangle$ denotes the sequence of proposed states. Clearly, the $m$'th proposal was accepted if $\boldsymbol{y}[m] = \boldsymbol{x}[m]$ and rejected otherwise. We denote these event using the indicators $S[m] = \boldsymbol{1}\{\boldsymbol{x}[m] = \boldsymbol{y}[m]\}$.

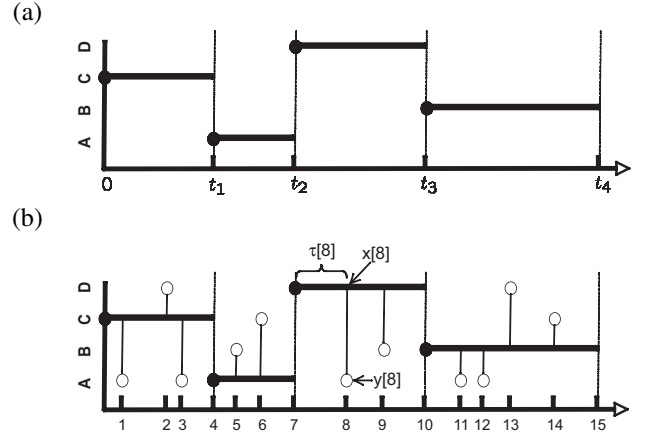

Figure 2: An illustration of training data. (a) A complete trajectory. The $x$-axis denotes time and the $y$-axis denotes the state at each time. Filled circles denote transitions. (b) A trajectory annotated with accepted and rejected proposals (closed and open circles, respectively). (Remember that accepted proposals lead to a transition.) The marks on the $x$-axis denote the index of the proposal. We illustrate the notation we use in the text, where $\tau[i]$ denotes the time interval before the $i$'th proposal, $\boldsymbol{x}[i]$ denote the actual state after the $i$'th proposal, and $\boldsymbol{y}[i]$ denote the proposed state in the $i$'th proposal.

The likelihood of these observations is the product of the probability density of the duration between proposals, and the probability of accepting or rejecting each proposal. Plugging in the factored form of $\boldsymbol{R}$ and $\boldsymbol{\pi}$ we can write this likelihood in a compact form.

**Proposition 6.1:** *Given an augmented data set, $\boldsymbol{\tau}$, $\Xi$, and $\Upsilon$, the log-likelihood can be decomposed as*

$$\ell(\boldsymbol{\theta}, \{\boldsymbol{R}^i\} : \boldsymbol{\tau}, \Xi, \Upsilon) = \sum_{i=1}^{n} \ell_{r,i}(\boldsymbol{R}^i : \boldsymbol{\tau}) + \ell_{\boldsymbol{s}}(\boldsymbol{\theta} : \Xi, \Upsilon),$$

*such that*

$$\ell_{r,i}(\boldsymbol{R}^i : \boldsymbol{\tau}) = \sum_{x_i \neq y_i} \left( M[x_i, y_i] \ln r^i_{x_i, y_i} - r^i_{x_i, y_i} T[x_i] \right)$$

*and*

$$\ell_s(\boldsymbol{\theta} : \Xi, \Upsilon) = \\ \sum_{i=1}^{n} \sum_{\boldsymbol{u}_i} \sum_{x_i \neq y_i} M^a[x_i, y_i | \boldsymbol{u}_i] \ln f(g_i(x_i \rightarrow y_i | \boldsymbol{u}_i)) + \\ \sum_{i=1}^{n} \sum_{\boldsymbol{u}_i} \sum_{x_i \neq y_i} M^r[x_i, y_i | \boldsymbol{u}_i] \ln(1 - f(g_i(x_i \rightarrow y_i | \boldsymbol{u}_i)))$$

where $M^a[x_i, y_i | \boldsymbol{u}_i]$ is the number of accepted transitions of $X_i$ from $x_i$ to $y_i$ when $\mathcal{N}(i) = \boldsymbol{u}_i$, $M^r[x_i, y_i | \boldsymbol{u}_i]$ is the count of rejected proposals to make the same transition, $M[x_i, y_i | \boldsymbol{u}_i] = M^a[x_i, y_i | \boldsymbol{u}_i] + M^r[x_i, y_i | \boldsymbol{u}_i]$, and $T[x_i]$ is the time spent in states where $X_i = x_i$.

Note that if we use $f_{\text{logistic}}$, then, as $\ln((1+e^{-x})^{-1})$ is concave, the likelihood function $\ell_s(\boldsymbol{\theta} : \Xi, \Upsilon)$ is concave and has a unique maximum.

### 6.2 Maximizing the Likelihood Function

Under the Maximum Likelihood Principle, our estimated parameters are the ones that maximize the likelihood function given the observations. We now examine how to maximize the likelihood. The decoupling of the likelihood into several terms allows us to estimate each set of parameters separately.

The estimation of $\boldsymbol{R}^i$ is straightforward: imposing the symmetry condition, the maximum likelihood estimate is

$$r^i_{x_i, y_i} = \frac{M[x_i, y_i] + M[y_i, x_i]}{T[x_i] + T[y_i]}.$$

Finding the maximum likelihood parameters of $\boldsymbol{\pi}$ is somewhat more involved. Note that the likelihood $\ell_s(\boldsymbol{\theta} : \Xi, \Upsilon)$ is quite different from the likelihood of a log-linear distribution given i.i.d. data [3]. The probability of acceptance or rejection involves ratios of probabilities. Therefore, the partition function $Z(\boldsymbol{\theta})$ cancels out, and does not appear in the likelihood.

In a sense, our likelihood is closely related to the *pseudo-likelihood* for log-linear models [1]. Recall that pseudo-likelihood is a technique for estimating the parameters of a Markov network (or log-linear model) that uses a different objective function. Rather than optimizing the joint likelihood, one optimizes a sum of log conditional likelihood terms, one for each variable given its neighbors. By considering the conditional probability of a variable given its neighbors, the partition function cancels out, allowing the parameters to be estimated without the use of inference. At the large sample limit, optimizing the pseudo-likelihood criterion is equivalent to optimizing the joint likelihood, but the results for finite sample sizes tend to be worse. In our setting, the generative model is defined in terms of ratios only. Thus, in this case the exact likelihood turns out to take a form similar to the pseudo-likelihood criterion. As for pseudo-likelihood, this form allows us to perform parameter estimation without requiring inference in the underlying Markov network.

In the absence of an analytical solution for this equation we learn the parameters using a gradient-based optimization procedure to find a (local) maximum of the likelihood. The derivation of the gradient is a standard exercise; for completeness, we provide the details in the appendix. When using $f_{\text{logistic}}$ we are guaranteed that such a procedure finds the unique global maximum.

### 6.3 Completing the Data

Our derivation of the likelihood and the associated optimization procedure relies on the assumption that rejected transition attempts are also observed in the data. As we can see from the form of the likelihood, these failures play an important role in estimating the parameters. The question is how to adapt the procedure to the case where rejected proposals are not observed. Our solution to this problem is to use Expectation Maximization, where we view the proposal attempts as the unobserved variables.

In this approach, we start with an initial guess of the model parameters. We use these to estimate the expected number of rejected proposals; we then treat these expected counts as though they were real, and maximize the likelihood using the procedure described in the previous section. We repeat these iterations until convergence.

The question is how to compute the expected number of rejected attempts. It turns out that this computation can be done analytically.

**Proposition 6.2:** *Given a CTMN, and an observed trajectory $\boldsymbol{\tau}, \Xi$. Then,*

$$\boldsymbol{E}[M^r[x_i, y_i | \boldsymbol{u}_i] | \mathcal{D}] \quad (9) \\ = T[x_i | \boldsymbol{u}_i] r^i_{x_i, y_i} (1 - f(g(x_i, y_i | \boldsymbol{u}_i)))$$

*where $T[x_i | \boldsymbol{u}_i]$ is the total amount of time the system was in states where $X_i = x_i$ and $\mathcal{N}(i) = \boldsymbol{u}_i$.*

We see that, in this case, the E-step of EM is fairly straightforward. The harder step is the M-step which requires an iterative gradient-based optimization procedure.

To summarize the procedure, to learn from complete data we perform the following steps: We first collect sufficient statistics $T[x_i | \boldsymbol{u}_i]$ and $M^a[x_i, y_i | \boldsymbol{u}_i]$. We then initialize the model with some set of parameters (randomly, or using prior knowledge). We then iterate over the two steps of EM until convergence: in the E-step, we complete the sufficient statistics with the expected number of rejected attempts, as per Eq. (9); in the M-step, we perform maximum likelihood estimation using the expected sufficient statistics, using gradient descent with the gradient of Eq. (10).

## 7 A Numerical Example

To illustrate the properties of our CTMN learning procedure, we evaluated it on a small synthetic data set. We

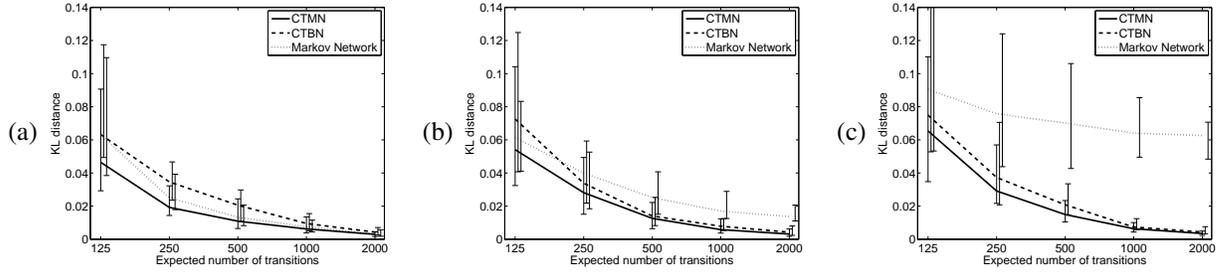

Correct structure

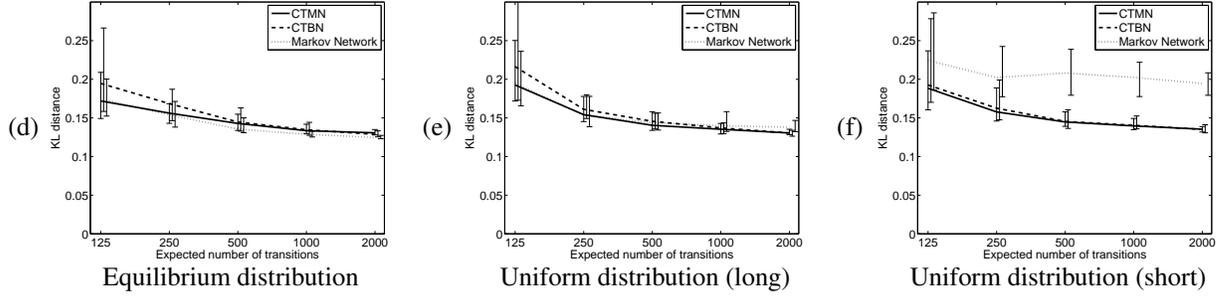

Partial structure

Equilibrium distribution     Uniform distribution (long)     Uniform distribution (short)

Figure 3: Comparison of estimates of the equilibrium distribution by the CTMN learning procedure (solid lines), the CTBN learning procedure (dashed lines) and a Markov Network parameter learning procedure applied to the frequency of time spent in each state (dotted lines). The $x$-axis denotes the total length of training trajectories (measured in units of expected number of observed transitions). The $y$-axis denotes the KL-divergence between the equilibrium distribution of the true model and the estimated model. The curves report the median performance among 50 data sets, and the error bars report $25\% - 75\%$ percentiles. (a-c) report performance when learning with the true structure from which the data was generated, and (d-f) report results when learning the parameters of a structure without the edges between $X_1$ and $X_4$. In (a) and (d) $p(\boldsymbol{X}(0))$ is the equilibrium distribution. In (b) and (e) $p(\boldsymbol{X}(0))$ is uniform and each trajectory is of length 25 time units. In (c) and (f) $p(\boldsymbol{X}(0))$ is uniform and each trajectory is of length 10 time units.

generated data from the CTMN model of Example 4.1 with $\boldsymbol{\theta} = \langle -0.2, -2.3, 0.7, 0.7, -1.2, -1.2, -1.2, -1.2 \rangle$ and proposal rates $r_{0,1}^1 = 1$, $r_{0,1}^2 = 2$, $r_{0,1}^3 = 3$, and $r_{0,1}^4 = 4$.

The goal of our experiments is to test the ability of the CTMN learning procedure to estimate stationary distributions from data in various conditions. As a benchmark, we compared our procedure to two alternative methods:

- A procedure that estimates the stationary distribution directly from the frequency of visiting each state. This procedure is essentially the standard parameter learning method for Markov networks, where the weight of each state (instance) is proportional to the duration in which the process spends in that state. This procedure uses gradient ascent to maximize the likelihood [3]. When the process is sampling from the stationary distribution, the relative time in each state is proportional to its stationary probability, and in such situations we expect this procedure to perform well.

- A procedure that estimates the $\boldsymbol{Q}$-matrix of the associated CTBN shown in Figure. 1. Here we used the methods designed for parameter learning of CTBNs in [9]. Once that the $\boldsymbol{Q}$-matrix has been estimated, the estimated stationary distribution is the only normalized vector in its null space.

We examined these three procedures in three sets of synthetic trajectories. The first set was generated by sampling the initial state $\boldsymbol{X}(0)$ of each trajectory from the stationary distribution and then sampling further states and durations from the target model. In this data set the system is in equilibrium throughout the trajectory. The second data set was generated by sampling the initial state from a uniform distribution, and so the system starts in a distribution that is far from equilibrium. However, the trajectory is long enough to let the system equilibrate. The third data set is similar to the second, except that trajectories are shorter and thus do not have sufficient time to equilibrate. To evaluate the effect of training set size, we repeated the learning experiments with different numbers of trajectories. We report the size of the training set in terms of the total length of training trajectories. Time is reported in units of *expected transition number*. That is, one time unit is equal to the average time between transitions when the process is in equilibrium. The short and long trajectories in our experiments are of length 10 and 25 expected transitions, respectively.

To evaluate the quality of the learned distribution, we measured the Kullback-Leibler divergences from the true stationary distribution to the estimated ones. Figures 3(a-c) show the results of these experiments. When sampling from the stationary distribution, the three procedures tend, as the data size increases, toward the correct distribution. For small data size, the performance of the CTMN learning procedure is consistently superior, although the error bars partially overlap. We start seeing a difference between the estimation procedures when we modify the initial distribution. As expected, the Markov network learning procedure suffers since it is learning from a biased sample. On the other hand, the performance of the CTMN and CTBN learning procedures is virtually unchanged, even when we modify the length of the trajectories. These results illustrate the ability of the CTMN and CTBN learning procedures to robustly estimate the equilibrium distribution from the dynamics even when the sampled process is not at equilibrium.

To test the robustness to the network structure, we also tested the performance of these procedures when estimating using a wrong structure. As we can see in Figures 3(d-f), while the three procedures converge to the wrong distribution, their relative behavior remains similar to the previous experiment, and the performance of the CTMN learning procedure is still not affected by the nature of the data.

## 8  Discussion and Future Work

In this paper, we define the framework of continuous time Markov networks, where we model a dynamical system as being governed by two factors: a local transition model, and a global acceptance/rejection model (based on an equilibrium distribution). By using a Markov network (or feature-based log-linear model) to encode the equilibrium distribution, we naturally define a temporal process guaranteed to have an equilibrium distribution of a particular, factored form. We showed a reduction from CTMNs to CTBNs that illustrates the differences in the expressive powers of the two formalisms. Moreover, this reduction allows us to reason in CTMNs by exploiting the efficient approximate inference algorithms for CTBNs. Finally, we provided learning algorithms for CTMNs, which allow us to learn the equilibrium distribution in a way that exploits our understanding about the system dynamics. We demonstrated on that this learning procedure is able to robustly estimate the equilibrium distribution even when the sampled process is not at equilibrium. These results can be combined for learning from partial observations, by plugging in the learning procedure as the M-step in the EM procedure for CTBNs [10].

This work opens many interesting questions. A key goal in learning these models is to estimate the stationary distribution. It is interesting to analyze, both theoretically and empirically, the benefit gained in this task by accounting for the process dynamics, as compared to learning the stationary distribution directly from a set of snapshots of the system (e.g., a set of instances of a protein sequence in different species). Moreover, so far, we have tackled only the problem of parameter estimation in these models. In many applications, the model structure is unknown, and of great interest. For example, in models of protein evolution, we want to know which pair of positions in the protein are directly correlated, and therefore likely to be structurally interacting. Of course, tackling this problem involves learning the structure of a Markov network, a notoriously difficult task. From the perspective of inference, our reduction to CTBNs can lose much of the structure of the model. For example, if the stationary distribution is a pairwise Markov network, the fact that the interaction model decomposes over pairs of variables is lost in the induced CTBN. It is interesting to see whether one can construct inference algorithms that better exploit this structure. Finally, one important limitation of the CTMN framework is the restriction to an exponential distribution on the duration between proposed state changes. Although such a model is a reasonable one in many systems (e.g., biological sequence evolution), there are other settings where it is too restrictive. In recent work, Nodelman et al. [10] show how one can expand the framework of CTBNs to allow a richer set of duration distributions. Essentially, their solution introduces a "hidden state" internal to a variable, so that the overall transition model of the variable is actually the aggregate of multiple transitions of its internal state. A similar solution can be applied in our setting, but the resulting model would not generally encode a reversible CTMP.

One major potential field of application for this class of models is sequence evolution. The current state of the art in phylogenetic inference is based on continuous time probabilistic models of evolution [4]. Virtually all of these models assume that sequence positions evolve independently of each other (although in some models, there are global parameters that induce weak dependencies). Our models provide a natural language for modeling such dependencies. In this domain, the proposal process corresponds to mutation rates within the sequence, and the equilibrium distribution is proportional to the relative fitness of different sequences. The latter function is of course very complex, but there is empirical evidence that modeling pairwise interactions can provide a good approximation [13]. Thus, in these systems, both the local mutation process and a factored equilibrium distribution are very appropriate, making CTMNs a potentially valuable tool for modeling and analysis. We hope to incorporate this formalism within phylogenetic inference tools, and to develop a methodology to leverage these models to provide new insights about the structure and function of proteins.

**Acknowledgments**

We thank A. Jaimovich, T. Kaplan, M. Ninio, I. Wiener, and the anonymous reviewers for comments on earlier versions of this manuscript. This work was supported by grants from the Israel Science Foundation and the US-Israel Binational Science Foundation, and by DARPA's CALO program, under sub-contract to SRI International.

## A  Gradient for Learning CTMNs

We now compute the derivative of the gradient of the log-likelihood, as specified in Proposition 6.1. The parameters $\boldsymbol{\theta}$ appear within the scope of the $g_i$ functions. Thus, to find the derivatives we differentiate these functions with respect to the parameters, and then apply the chain rule for derivatives:

$$\frac{\partial}{\partial \theta_k} \ell_{\boldsymbol{s}}(\boldsymbol{\theta} : \Xi, \Upsilon) = \quad (10)$$

$$\sum_{i: X_i \in \boldsymbol{D}_k} \sum_{\boldsymbol{u}_i} \sum_{x_i \neq y_i} \Delta_k(x_i, y_i | \boldsymbol{u}_i) (\psi_a(x_i, y_i | \boldsymbol{u}_i) M^a [x_i, y_i | \boldsymbol{u}_i] - \psi_r(x_i, y_i | \boldsymbol{u}_i) M^r [x_i, y_i | \boldsymbol{u}_i])$$

where

$$\begin{aligned}
\Delta_k(x_i, y_i | \boldsymbol{u}_i) &= s_k(\boldsymbol{u}_i, y_i) - s_k(\boldsymbol{u}_i, x_i) \\
\psi_a(x_i, y_i | \boldsymbol{u}_i) &= \left. \frac{z f'(z)}{f(z)} \right|_{z = g_i(x_i \to y_i | \boldsymbol{u}_i)} \\
\psi_r(x_i, y_i | \boldsymbol{u}_i) &= \left. \frac{z f'(z)}{1 - f(z)} \right|_{z = g_i(x_i \to y_i | \boldsymbol{u}_i)}
\end{aligned}$$

This shows that the update of $\theta_k$ is a weighted combination of the contribution of each proposed transition. The weight of the transition depends on how sensitive the ratio of probabilities is to the feature, denoted by $\Delta_k(x_i, y_i | \boldsymbol{u}_i)$ and the number of times this transition was accepted or rejected, captured by the empirical counts. In addition, each proposal is weighted by $\psi_a(x_i, y_i | \boldsymbol{u}_i)$, which captures the improbability of the acceptance (respectively rejection for $\psi_r(x_i, y_i | \boldsymbol{u}_i)$). The less probable they are, the larger the change in $\theta_k$.

We can get better understanding of these terms if we consider their value for specific choices of $f$. For example, if we use $f_{\text{logistic}}$, then

$$\begin{aligned}
\psi_a(x_i, y_i | \boldsymbol{u}_i) &= 1 - f_{\text{logistic}}(g_i(x_i \to y_i | \boldsymbol{u}_i)) \\
\psi_r(x_i, y_i | \boldsymbol{u}_i) &= f_{\text{logistic}}(g_i(x_i \to y_i | \boldsymbol{u}_i)),
\end{aligned}$$

that is, the rejection and acceptance probabilities, respectively. The smaller these values, the more probable was the acceptance (resp. rejection) and so it results in a smaller gradient of the likelihood in the direction of this parameter. When using $f_{\text{Metropolis}}$ the two functions are not symmetric:

$$\begin{aligned}
\psi_a(x_i, y_i | \boldsymbol{u}_i) &= \boldsymbol{I}\{g_i(x_i \to y_i | \boldsymbol{u}_i) < 1\} \\
\psi_r(x_i, y_i | \boldsymbol{u}_i) &= \\
& \boldsymbol{I}\{g_i(x_i \to y_i | \boldsymbol{u}_i) > 1\} f_{\text{logistic}}(g_i(x_i \to y_i | \boldsymbol{u}_i))
\end{aligned}$$

with a discontinuity when $g_i(x_i \to y_i | \boldsymbol{u}_i) = 1$. We see that, in this case, the updates are asymmetric, with maximal weight to updates of accepted transitions.

## References

[1] J. Besag. On the statistical analysis of dirty pictures. *J. Roy. Stat. Soc. B Met.*, 48(3):259–302, 1986.

[2] K.L. Chung. *Markov chains with stationary transition probabilities*. 1960.

[3] S. Della Pietra, V. Della Pietra, and J. Lafferty. Inducing features of random fields. *IEEE Trans. PAMI*, 19(4):380–393, 1997.

[4] J. Felsenstein. *Inferring Phylogenies*. 2004.

[5] D. Heckerman, D. Geiger, and D. M. Chickering. Learning Bayesian networks: The combination of knowledge and statistical data. *Mach. Learn.*, 20:197–243, 1995.

[6] N. Metropolis, A.W. Rosenbluth, M.N. Rosenbluth, A.H. Teller, and E. Teller. Equation of state calculation by fast computing machines. *J. Chem. Phys.*, 21:1087–1092, 1953.

[7] B. Ng, A. Pfeffer, and R. Dearden. Continuous time particle filtering. In *IJCAI '05*. 2005.

[8] U. Nodelman, C.R. Shelton, and D. Koller. Continuous time Bayesian networks. In *UAI '02*, pp. 378–387. 2002.

[9] U. Nodelman, C.R. Shelton, and D. Koller. Learning continuous time Bayesian networks. In *UAI '03*, pp. 451–458. 2003.

[10] U. Nodelman, C.R. Shelton, and D. Koller. Expectation maximization and complex duration distributions for continuous time Bayesian networks. In *UAI '05*, pp. 421–430. 2005.

[11] U. Nodelman, C.R. Shelton, and D. Koller. Expectation propagation for continuous time Bayesian networks. In *UAI '05*, pp. 431–440. 2005.

[12] J. Pearl. *Probabilistic Reasoning in Intelligent Systems*. 1988.

[13] M. Socolich, *et al.* Evolutionary information for specifying a protein fold. *Nature*, 437(7058):512–518, 2005.

[14] H.M. Taylor and S. Karlin. *An Introduction to Stochastic Modeling*. 1998.